\documentclass{article}

\usepackage{microtype}
\usepackage{graphicx}
\usepackage{subcaption}
\usepackage{booktabs} 
\usepackage{xcolor}
\usepackage{colortbl} 
\usepackage{multirow} 

\usepackage{hyperref}

\usepackage{enumitem}

\usepackage[preprint]{icml2026}

\usepackage{amsmath}
\usepackage{amssymb}
\usepackage{mathtools}
\usepackage{amsthm}

\usepackage[capitalize,noabbrev]{cleveref}

\usepackage{pgfplots}
\pgfplotsset{compat=1.18}
\usepgfplotslibrary{groupplots,fillbetween}
\usetikzlibrary{patterns,positioning,shadows,decorations.pathreplacing}

\definecolor{ExpColor}{RGB}{76,114,176}
\definecolor{RevColor}{RGB}{221,132,82}
\definecolor{SuccessColor}{RGB}{85,168,104}
\definecolor{FailColor}{RGB}{196,78,82}
\definecolor{NeutralColor}{RGB}{129,114,179}
\definecolor{AdvColor}{RGB}{166,86,40}

\usepackage[capitalize,noabbrev]{cleveref}

\theoremstyle{plain}

\theoremstyle{definition}

\theoremstyle{remark}

\usepackage{xspace}
\usepackage{fontawesome5}  
\newcommand{\eg}{\emph{e.g.}\xspace}

\usepackage[textsize=tiny]{todonotes}

\usepackage[most]{tcolorbox}

\definecolor{ExcerptBorder}{RGB}{120,120,120}
\definecolor{ExplorationBG}{RGB}{245,245,245}
\definecolor{ReviewBG}{RGB}{233,242,252}

\definecolor{StudentRed}{RGB}{180,50,50}
\definecolor{StudentBG}{RGB}{252,235,235}
\definecolor{TeacherGreen}{RGB}{60,120,60}
\definecolor{TeacherBG}{RGB}{235,248,235}

\definecolor{PreExecBlue}{RGB}{50,130,180}
\definecolor{PreExecBG}{RGB}{232,244,252}

\definecolor{MidExecPurple}{RGB}{95,75,140}
\definecolor{MidExecBG}{RGB}{244,240,252}

\definecolor{PostExecGreen}{RGB}{90,140,110}
\definecolor{PostExecBG}{RGB}{240,248,242}

\definecolor{AdvCoral}{RGB}{200,85,60}
\definecolor{AdvBG}{RGB}{255,242,238}

\newtcolorbox{promptbox}[3][]{%
  enhanced,
  colback=#3,
  colframe=#2,
  colbacktitle=#2,
  coltitle=white,
  fonttitle=\bfseries\sffamily,
  title=#1,
  boxrule=0.6pt,
  leftrule=5pt,
  titlerule=0pt,
  arc=5pt,
  outer arc=5pt,
  fuzzy shadow={3pt}{-3pt}{0pt}{0.5pt}{black!40},
  top=6pt,
  bottom=6pt,
  left=8pt,
  right=8pt,
  toptitle=5pt,
  bottomtitle=5pt,
  fontupper=\small\ttfamily,
  before upper={\setlength{\parskip}{5pt}},
  height=6.5cm,
  space to lower,
}

\newcommand{\preexecbox}[2]{%
  \begin{promptbox}[#1]{PreExecBlue}{PreExecBG}#2\end{promptbox}%
}
\newcommand{\midexecbox}[2]{%
  \begin{promptbox}[#1]{MidExecPurple}{MidExecBG}#2\end{promptbox}%
}
\newcommand{\postexecbox}[2]{%
  \begin{promptbox}[#1]{PostExecGreen}{PostExecBG}#2\end{promptbox}%
}
\newcommand{\advpostbox}[2]{%
  \begin{promptbox}[#1]{AdvCoral}{AdvBG}#2\end{promptbox}%
}

\newtcolorbox{agentquote}[3][]{%
  enhanced,
  colback=#3,
  colframe=#2,
  boxrule=0.4pt,
  leftrule=4pt,
  arc=3pt,
  outer arc=3pt,
  top=4pt,
  bottom=4pt,
  left=6pt,
  right=6pt,
  fontupper=\small,
  before upper={\setlength{\parskip}{3pt}},
  #1
}

\newcommand{\prequote}[2]{%
  \begin{agentquote}{PreExecBlue}{PreExecBG}%
    \textbf{\sffamily\textcolor{PreExecBlue}{Pre-Exec.} \hfill #1}\\[2pt]%
    \ttfamily #2%
  \end{agentquote}%
}
\newcommand{\postquote}[2]{%
  \begin{agentquote}{PostExecGreen}{PostExecBG}%
    \textbf{\sffamily\textcolor{PostExecGreen}{Post-Exec.} \hfill #1}\\[2pt]%
    \ttfamily #2%
  \end{agentquote}%
}
\newcommand{\advquote}[2]{%
  \begin{agentquote}{AdvCoral}{AdvBG}%
    \textbf{\sffamily\textcolor{AdvCoral}{Adv. Post-Exec.} \hfill #1}\\[2pt]%
    \ttfamily #2%
  \end{agentquote}%
}
\newcommand{\midquote}[2]{%
  \begin{agentquote}{MidExecPurple}{MidExecBG}%
    \textbf{\sffamily\textcolor{MidExecPurple}{Mid-Exec.} \hfill #1}\\[2pt]%
    \ttfamily #2%
  \end{agentquote}%
}

\icmltitlerunning{Agentic Uncertainty Reveals Agentic Overconfidence}

\begin{document}

\twocolumn[
  \icmltitle{Agentic Uncertainty Reveals Agentic Overconfidence}



  \icmlsetsymbol{equal}{*}

  \begin{icmlauthorlist}
    \icmlauthor{Jean Kaddour}{ucl,sevn}
    \icmlauthor{Srijan Patel}{sevn}
    \icmlauthor{Gbètondji Dovonon}{ucl}
    \icmlauthor{Leo Richter}{ucl}
    \icmlauthor{Pasquale Minervini}{ed}
    \icmlauthor{Matt J. Kusner}{mila,poly}
  \end{icmlauthorlist}

  \icmlaffiliation{ucl}{University College London}
  \icmlaffiliation{sevn}{SevnAI}
  \icmlaffiliation{ed}{University of Edinburgh}
\icmlaffiliation{mila}{Mila - Québec AI Institute}
\icmlaffiliation{poly}{Polytechnique Montréal}
  \icmlcorrespondingauthor{Jean Kaddour}{jean.kaddour.20@ucl.ac.uk}

  \icmlkeywords{Machine Learning, ICML}

  \vskip 0.3in
]



\printAffiliationsAndNotice{Code available at \url{https://github.com/sevn-ai/agentic-uncertainty}.}

\begin{abstract}
Can AI agents predict whether they will succeed at a task? We study \textbf{agentic uncertainty} by eliciting success probability estimates before, during, and after task execution. All results exhibit \textbf{agentic overconfidence}: some agents that succeed only 22\% of the time predict 77\% success. Counterintuitively, pre-execution assessment with strictly less information \emph{tends} to yield better discrimination than standard post-execution review, though differences are not always significant. Adversarial prompting reframing assessment as bug-finding achieves the best calibration.
\end{abstract}

\section{Introduction}

A software engineer needs to fix an auth service error. Before delegating to an AI coding agent, she asks: what are the chances this succeeds?

\prequote{P(success): 72\%}{The issue is clear and well-defined, the error message points directly to the problem, and the fix follows existing patterns in the codebase.}

\begin{figure}[!t]
\centering
\includegraphics[width=\columnwidth]{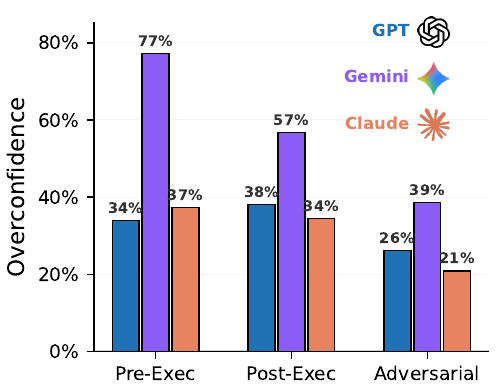}
\caption{\textbf{Agentic overconfidence}. We measure the overconfidence as the difference between the estimated success probability and the true success probability (true rates: GPT-5.2 Codex 35\%, Gemini-3-Pro 22\%, Opus 4.5 27\%). We plot three strategies: pre-, post-, and adversarial-post-execution. All agents systematically overestimate their success.}
\label{fig:hero_results}
\end{figure}

72\% confidence before any code is written. As the coding agent works, she asks another agent to monitor progress:

\midquote{P(success): 78\%}{The agent has correctly diagnosed the problem and knows exactly what code to add. The probability of success is high.}

Confidence \emph{rises} to 78\%. The patch is now complete. She fires off a review agent:

\postquote{P(success): 92\%}{The patch is a correct and complete fix. It's a minimal, focused change that adds the missing interface method.}

Too optimistic. Let's spawn an adversarial agent.

\advquote{P(success): 85\%}{Minor concerns don't affect the main use case... the patch correctly resolves the reported issue.}

Still 85\%. All four agents confidently predict success.

But the patch fails! And this \textbf{agentic overconfidence} is systematic. For example, GPT-5.2-Codex-based post-execution agents predict 73\% success against a true rate of 35\% averaged over 100 SWE-Bench-Pro \cite{deng2025swebenchpro} tasks.

This matters because the scope of autonomous work is expanding rapidly. The effective length of tasks that AI agents complete has doubled every 7 months for six years~\cite{measuring-ai-ability-to-complete-long-tasks}. As we increasingly delegate complex workflows to agents
\cite{appel2025anthropiceconomicindexreport}, we must develop scalable oversight protocols \cite{bowman2022measuringprogressscalableoversight}.

In this work, we elicit \emph{agentic uncertainty} at three points in a coding agent's lifecycle: pre-, mid-, and post-execution. Each corresponds to a different oversight question: Can agents predict failure before committing resources? Can they recognize failure as it unfolds? Can they verify their own work? Importantly, we use the same underlying model for both the coding agent that produces patches and the uncertainty agent that assesses them, isolating the effect of information access from differences in model capability.

Our experiments on 100 SWE-bench Pro tasks across three frontier models (GPT-5.2-Codex, Gemini-3-Pro, Claude Opus 4.5) reveal several striking findings:
\begin{itemize}[leftmargin=*, itemsep=1pt, topsep=2pt]
    \item \textbf{Pervasive overconfidence.} Post-execution agents can predict 73\% success on average against a 35\% base rate (GPT), with similar gaps across all models.
    \item \textbf{More context, uncalibrated doubt.} Mid-execution agents develop ``cold feet'': confidence decreases as they observe their partial work, but this doubt is uninformative, occurring equally for successes and failures.
    \item \textbf{Adversarial framing helps.} Prompting agents to ``find bugs'' rather than ``verify correctness'' reduces overconfidence by up to 15 pp and tends to achieve the best calibration across models in our setup.
\end{itemize}

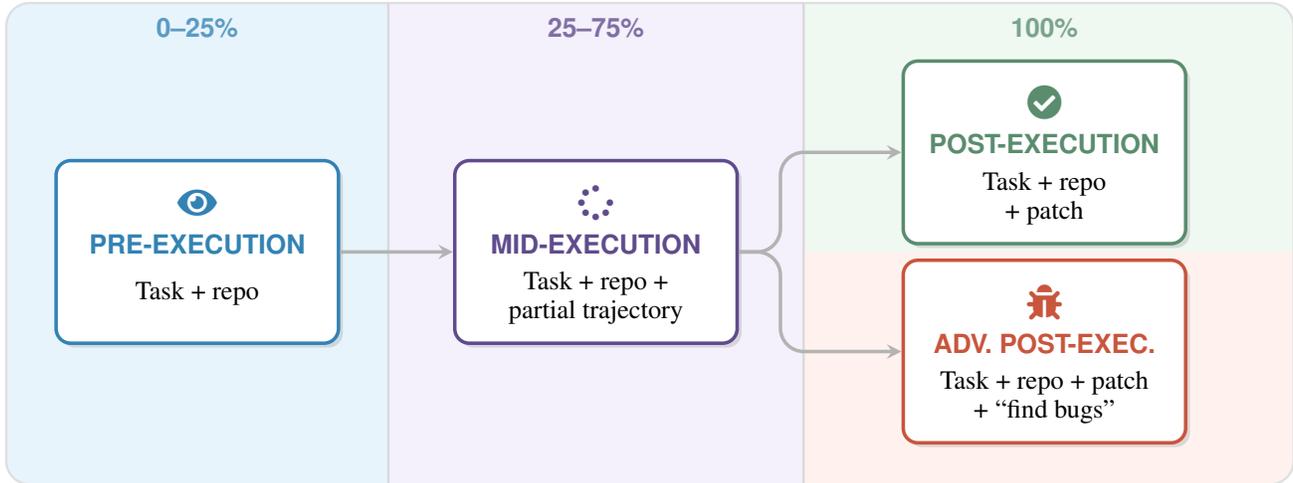
\begin{figure*}[t]
\centering
\resizebox{\textwidth}{!}{%
\begin{tikzpicture}[
    box/.style={draw, rounded corners=5pt, minimum width=3.4cm, minimum height=2.2cm, align=center},
    shadowbox/.style={box, drop shadow={shadow xshift=1.5pt, shadow yshift=-1.5pt, opacity=0.25}},
    prebox/.style={shadowbox, fill=white, draw=PreExecBlue, line width=1.2pt},
    midbox/.style={shadowbox, fill=white, draw=MidExecPurple, line width=1.2pt},
    postbox/.style={shadowbox, fill=white, draw=PostExecGreen, line width=1.2pt},
    advbox/.style={shadowbox, fill=white, draw=AdvCoral, line width=1.2pt},
    label/.style={font=\bfseries\sffamily\small},
    icon/.style={font=\large},
    arrow/.style={->, >=stealth, line width=1.2pt, gray!60},
]

\def\phasebot{-2.8}
\def\phasetop{3.0}
\def\preboundary{4.3}
\def\midboundary{9.3}
\def\postend{15.2}

\def\phaseround{8pt}
\begin{scope}
    \clip[rounded corners=\phaseround] (-0.3, \phasebot) rectangle (\postend, \phasetop);
    \fill[PreExecBG] (-0.3, \phasebot) rectangle (\preboundary, \phasetop);
    \fill[MidExecBG] (\preboundary, \phasebot) rectangle (\midboundary, \phasetop);
    \fill[PostExecBG] (\midboundary, 0) rectangle (\postend, \phasetop);
    \fill[AdvBG] (\midboundary, \phasebot) rectangle (\postend, 0);
\end{scope}
\draw[gray!25, line width=0.8pt, rounded corners=\phaseround] (-0.3, \phasebot) rectangle (\postend, \phasetop);

\draw[gray!30, line width=0.8pt] (\preboundary, \phasebot) -- (\preboundary, \phasetop);
\draw[gray!30, line width=0.8pt] (\midboundary, \phasebot) -- (\midboundary, \phasetop);

\node[font=\small\bfseries\sffamily, text=PreExecBlue!80] at (2, \phasetop-0.3) {0--25\%};
\node[font=\small\bfseries\sffamily, text=MidExecPurple!80] at (6.8, \phasetop-0.3) {25--75\%};
\node[font=\small\bfseries\sffamily, text=PostExecGreen!80] at (12.2, \phasetop-0.3) {100\%};

\node[prebox] (pre) at (2, 0) {};
\node[icon, text=PreExecBlue] at (2, 0.6) {\faEye};
\node[label, text=PreExecBlue] at (2, 0.1) {PRE-EXECUTION};
\node[font=\small, align=center] at (2, -0.5) {Task + repo};

\node[midbox] (mid) at (6.8, 0) {};
\node[icon, text=MidExecPurple] at (6.8, 0.6) {\faSpinner};
\node[label, text=MidExecPurple] at (6.8, 0.1) {MID-EXECUTION};
\node[font=\small, align=center] at (6.8, -0.55) {Task + repo +\\partial trajectory};

\node[postbox] (post) at (12.2, 1.2) {};
\node[icon, text=PostExecGreen] at (12.2, 1.8) {\faCheckCircle};
\node[label, text=PostExecGreen] at (12.2, 1.3) {POST-EXECUTION};
\node[font=\small, align=center] at (12.2, 0.65) {Task + repo\\+ patch};

\node[advbox] (adv) at (12.2, -1.2) {};
\node[icon, text=AdvCoral] at (12.2, -0.6) {\faBug};
\node[label, text=AdvCoral, align=center] at (12.2, -1.1) {ADV.\ POST-EXEC.};
\node[font=\small, align=center] at (12.2, -1.75) {Task + repo + patch\\+ ``find bugs''};

\draw[arrow] (pre.east) -- (mid.west);
\draw[arrow, rounded corners=8pt] (mid.east) -- ++(0.5,0) |- (post.west);
\draw[arrow, rounded corners=8pt] (mid.east) -- ++(0.5,0) |- (adv.west);

\end{tikzpicture}%
}
\caption{\textbf{Agentic Uncertainty Regimes.} Each regime observes different information. Post-execution and adversarial post-execution occur at the same point but use different prompts.}
\label{fig:overall}
\end{figure*}

\section{Methods}\label{sec:methods}
\subsection{Problem Setup}
We define \textbf{agentic uncertainty} as an agent's estimate of the probability that an agent built on the same underlying model will successfully complete a task. The uncertainty (-estimating) agent may use a different system prompt or have access to different information than the task-solving agent, but shares the same base model.

Unlike standard uncertainty quantification, which focuses on confidence in individual predictions or token probabilities, agentic uncertainty concerns the outcome of an entire multi-step trajectory: will this sequence of observations, reasoning, and actions culminate in task success?

\citet{kadavath2022language} introduced P(IK): ``probability that I know,'' measuring whether models can predict which questions they can answer correctly. We generalize this to agentic settings and call it \textbf{P(IS)}: ``probability that I succeed.'' Formally, given a task $t$, base model $M$, and information state $\mathcal{I}$ available at elicitation time:
\begin{equation}
  P(\text{IS}) \;\coloneqq\; P\!\left(\text{agent}_{M}\text{ succeeds on }t \;\middle|\; \mathcal{I}\right)
  \label{eq:pis}
\end{equation}
where $\mathcal{I}$ may include the task description, repository state, partial trajectory, or proposed patch, depending on the elicitation regime.
For example, where P(IK) asks ``do I know the capital of France?'', a question about factual recall, P(IS) asks ``can I fix this bug through a sequence of file edits, shell commands, and test runs?''

\subsection{Uncertainty Agents}
We consider three points in an agent's life cycle to elicit uncertainty estimates: before attempting a solution, during execution, and after producing a solution.

Each uncertainty agent interacts with a sandbox environment through tool use: reading files, searching code, and inspecting version history. This shared infrastructure means agents can actively gather evidence to inform their estimates. The key distinction is the information available at uncertainty estimation time.

\subsubsection{Pre-Execution Agent}
In the pre-execution setting, an agent receives only the task description (\eg, a GitHub issue) and read-only access to the repository. Crucially, the agent cannot execute code, run tests, or modify files. If it could attempt solutions and observe test results, it might anchor on its own partial solution rather than reasoning abstractly about the task.
The agent must form an estimate based on cues, e.g., the complexity of the codebase, rather than through trial and error.

\subsubsection{Mid-Execution Agent}
Mid-execution asks whether can we detect failure \emph{in progress}?
Rather than eliciting estimates at a single point, we prompt the agent at 25\%, 50\%, and 75\% of total agent steps (tool calls and reasoning turns) to estimate the success probability given the trajectory so far.
If these estimates reliably decrease as agents approach failure, we could implement early stopping or human escalation before resources are wasted on doomed approaches.

\subsubsection{Post-Execution Agent}
Once a task solution has been proposed, can we trust an agent's self-assessment that it is correct? This is the default scenario when ground-truth verification is unavailable. If post-execution agents are well-calibrated, we can use their success estimates to route submissions with low estimate of success to human reviewers while auto-accepting solutions with high estimates. If they are poorly calibrated, this delegation becomes dangerous.

In the post-execution setting, an agent receives both the task description and a proposed patch written by another agent. The repository is in its post-patch state, where the changes have already been applied, and the agent can explore the modified codebase. After assessment, the agent estimates whether the patch successfully solves the task.

\paragraph{Adversarial post-execution variant.} We also evaluate a variant that explicitly prompts agents to find bugs before estimating confidence. Rather than asking ``is this correct?'', adversarial post-execution asks ``what bugs can you find?'' This reframes the task from verification to falsification, potentially counteracting confirmation bias that is encouraged by the vanilla post-execution framing, asking whether a patch is correct.

\begin{figure*}[t]
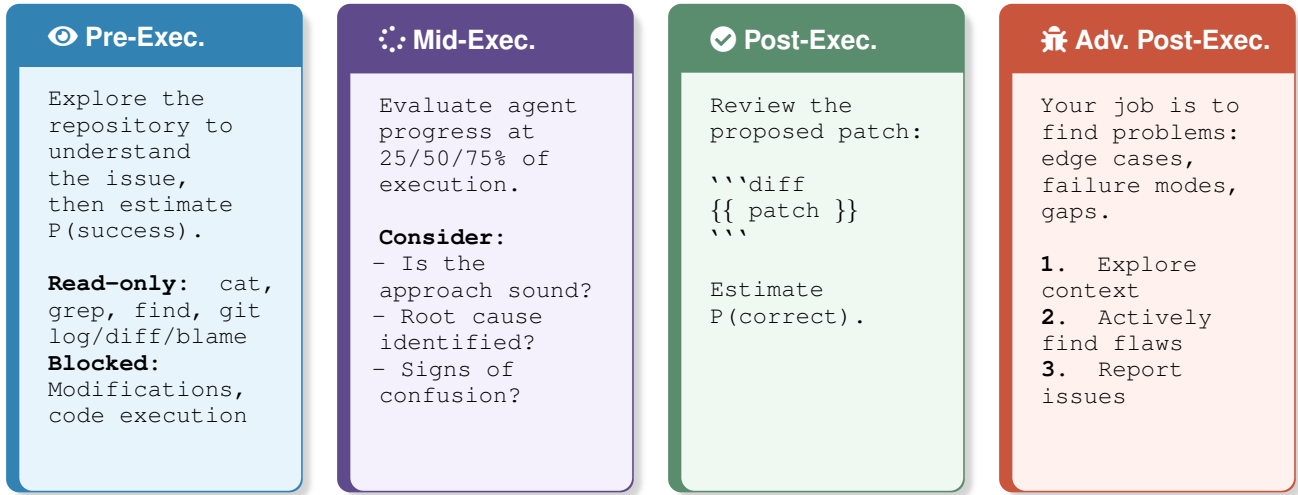

\centering
\begin{minipage}[t]{0.23\linewidth}
\preexecbox{\faEye\ Pre-Exec.}{Explore the repository to understand the issue, then estimate P(success).\newline\newline
\textbf{Read-only:} cat, grep, find, git log/diff/blame\newline
\textbf{Blocked:} Modifications, code execution}
\end{minipage}\hfill
\begin{minipage}[t]{0.23\linewidth}
\midexecbox{\faSpinner\ Mid-Exec.}{Evaluate agent progress at 25/50/75\% of execution.\newline\newline
\textbf{Consider:}\newline
- Is the approach sound?\newline
- Root cause identified?\newline
- Signs of confusion?}
\end{minipage}\hfill
\begin{minipage}[t]{0.23\linewidth}
\postexecbox{\faCheckCircle\ Post-Exec.}{Review the proposed patch:\newline\newline
\texttt{```diff}\newline
\{\{ patch \}\}\newline
\texttt{```}\newline\newline
Estimate P(correct).}
\end{minipage}\hfill
\begin{minipage}[t]{0.23\linewidth}
\advpostbox{\faBug\ Adv.\ Post-Exec.}{Your job is to find problems: edge cases, failure modes, gaps.\newline\newline
\textbf{1.} Explore context \newline
\textbf{2.} Actively find flaws\newline
\textbf{3.} Report issues \newline\newline}
\end{minipage}
\caption{\textbf{Uncertainty Agent Prompt Excerpts.} \emph{Pre-execution} explores the codebase before any solution attempt. \emph{Mid-execution} evaluates an agent's partial trajectory for signs of progress or struggle. \emph{Post-execution} reviews a proposed patch. \emph{Adversarial post-execution} explicitly prompts bug-finding before estimation. All agents output probability estimates $[0,100]$.}
\label{fig:prompts}
\end{figure*}

\section{Experiments}\label{sec:experiments}

\subsection{Setup}

We evaluate on 100 tasks sampled at random from SWE-bench Pro \citep{deng2025swebenchpro}, which requires substantial multi-file modifications (mean 107 lines across 4.1 files) where frontier models achieve only 23--44\% success. Each task corresponds to a full agentic trajectory, which can run up to roughly 15 minutes of wall-clock execution. We generate task-solving trajectories using GPT-5.2-Codex, Gemini 3 Pro, and Claude Opus 4.5, then evaluate uncertainty estimates from the same models. All uncertainty agents are implemented using mini-swe-agent\footnote{\url{https://github.com/SWE-agent/mini-swe-agent}} with read-only access to prevent ``peeking'' at test results. Figure~\ref{fig:prompts} shows prompt excerpts.

We measure \emph{discrimination} via AUROC (can agents distinguish successes from failures?). We measure \emph{calibration} via ECE, Brier score, and overconfidence (mean estimate minus base rate).

\subsection{Pervasive Overconfidence}

\begin{figure*}[t]
\centering
\includegraphics[width=\textwidth]{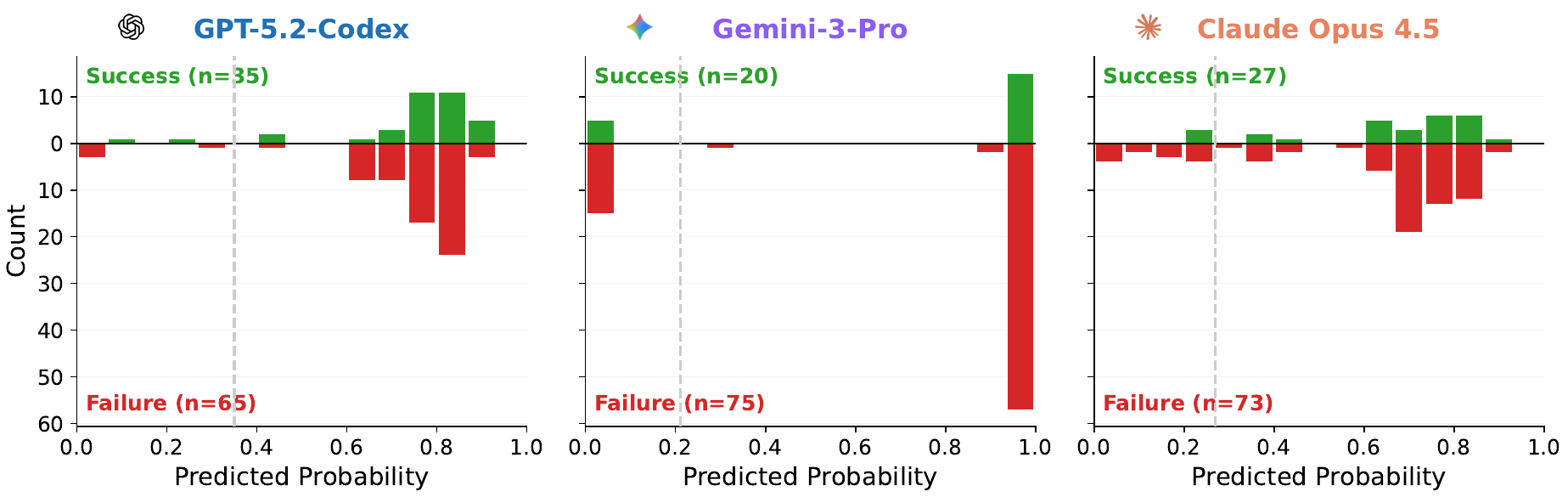}
\caption{\textbf{Distribution of post-execution confidence estimates by model.} Success cases shown above the axis (green), failure cases below (red); dashed lines indicate base rates. \emph{Mirror symmetry reveals indistinguishable distributions}: where bars match above and below, the model assigns identical confidence regardless of outcome. Gemini exhibits the most extreme pattern: nearly all predictions cluster at 100\% confidence, creating dramatic mirrored towers. This visual symmetry directly explains the poor discrimination: high-confidence predictions provide no signal about actual success.}
\label{fig:confidence_histograms}
\end{figure*}

Table~\ref{tab:unified_results} reveals systematic overconfidence across all models and methods. Post-execution agents predict 73\% success for GPT (base rate 35\%), 77\% for Gemini (base rate 22\%), and 61\% for Claude (base rate 27\%). Figure~\ref{fig:confidence_histograms} visualizes this through confidence distributions: both successes and failures cluster at high values, with near-complete overlap. Gemini exhibits the most extreme pattern, with predictions clustering near 100\% regardless of outcome. In fact, Gemini's pre-execution estimates average 99\%, leaving virtually no room to distinguish tasks by predicted difficulty. This suggests a distinct failure mode, reluctance to express uncertainty, beyond miscalibration.

\begin{figure*}[t]
\centering
\includegraphics[width=\textwidth]{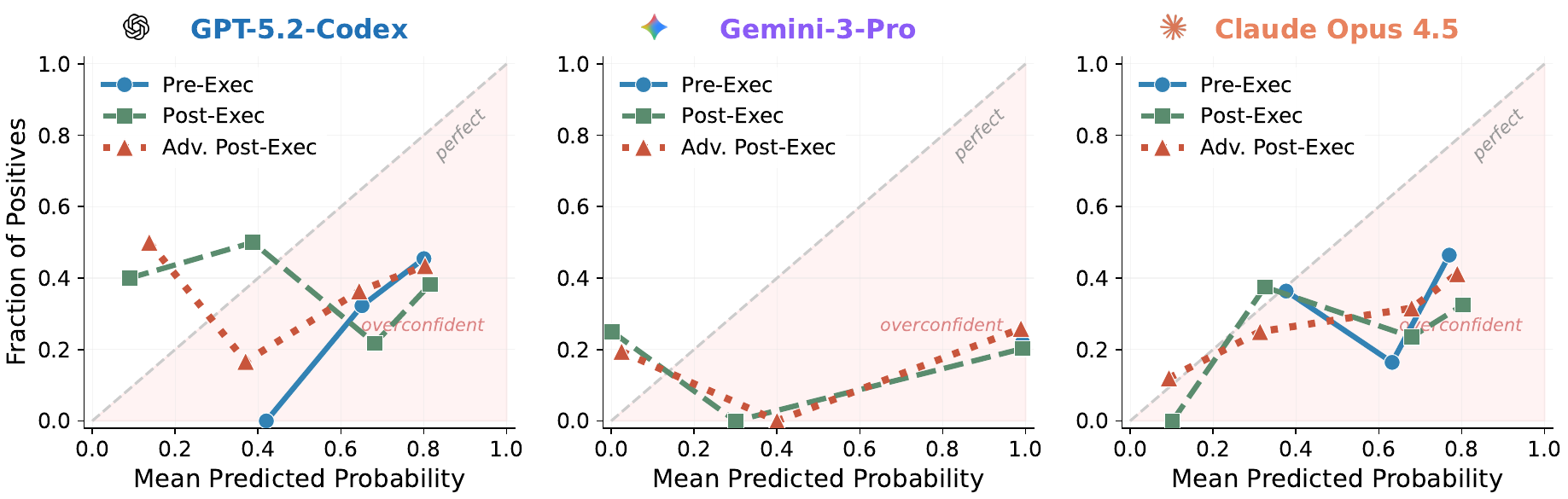}
\caption{\textbf{Calibration curves reveal systematic overconfidence.} Points below the diagonal (shaded region) indicate overconfidence: models predict higher success probability than achieved. All methods fall in this region across all models. Gemini shows the most severe miscalibration: predictions near 100\% yield only $\sim$20\% accuracy. The adversarial method (triangles) consistently shifts curves upward toward the diagonal, achieving the best calibration, while pre-execution (circles) shows less extreme overconfidence than standard post-execution (squares) for GPT and Claude.}
\label{fig:calibration_curves}
\end{figure*}

Figure~\ref{fig:calibration_curves} confirms miscalibration: all curves fall substantially below the diagonal, meaning 80\% confidence corresponds to far less than 80\% actual success. Adversarial post-execution (triangles) consistently achieves the best calibration across models.

This overconfidence is strikingly asymmetric. Across all models and methods, 62\% of predictions on failing instances are overconfident (predicted $\geq 0.7$), while only 11\% of predictions on passing instances are underconfident (predicted $< 0.3$). Agents are 5.5$\times$ more likely to confidently predict success on a failing task than to doubt a successful one. Adversarial prompting partially mitigates this: the overconfident-failure rate drops from 72\% (standard review) to 45\% (adversarial review).

\subsection{Less Information, Better Discrimination}\label{sec:less_info}

Pre-execution \emph{tends} to achieve better discrimination than vanilla post-execution across all three models despite having strictly less information: GPT 0.62 vs.\ 0.58, Claude 0.64 vs.\ 0.55, and Gemini 0.53 vs.\ 0.51.
Bootstrap 95\% confidence intervals are wide: GPT pre-execution AUROC is 0.62 [0.50, 0.75] versus post-execution 0.58 [0.46, 0.70]; Claude shows 0.64 [0.49, 0.78] versus 0.55 [0.42, 0.67]. So individual pairwise differences are not statistically significant. However, the consistency of the pattern across all three models (3--9~pp improvement each time) is suggestive.

One explanation is that observing a proposed patch anchors agents on surface plausibility rather than forcing abstract reasoning about task difficulty. Supporting this interpretation, adversarial post-execution, which counteracts such anchoring through bug-finding framing, recovers discrimination competitive with pre-execution for Claude (0.64 [0.51, 0.76]) and Gemini (0.57 [0.43, 0.70]), though not for GPT (0.55 [0.42, 0.68]). This pattern is consistent under AUPRC, which better handles class imbalance: GPT pre-execution achieves 0.52 versus post-execution's 0.42. Gemini's AUPRC remains poor across all methods (0.21--0.27), consistent with its near-chance AUROC.

A concrete example illustrates this anchoring effect. On a task requiring exposure of a new QtWebEngine dark-mode policy in \texttt{qutebrowser}, the pre-execution agent correctly identified the core risk:

\prequote{P(success): 45\%}{Biggest risk is knowing exact Chromium key/enum for classifier without external reference.}

After seeing the patch, the post-execution agent was reassured by the plausible-looking implementation and jumped to 82\%:

\postquote{P(success): 82\%}{Adds smart-simple value, Qt 6.6 variant gating, optional mapping to suppress switches{\ldots} logic seems sound.}

The patch failed. The pre-execution agent's abstract reasoning about task difficulty was more informative than the post-execution agent's assessment of a coherent-looking but incorrect solution.

\subsection{Mid-Execution: Uninformative Doubt}

\begin{figure}[t]
\centering
\includegraphics[width=\columnwidth]{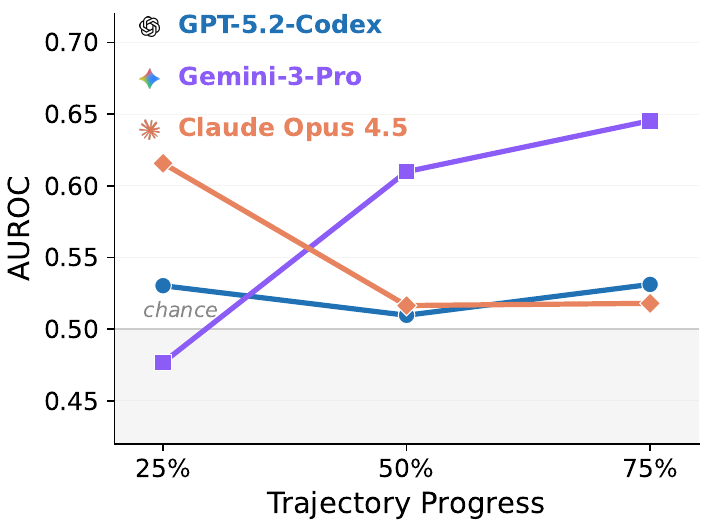}
\caption{\textbf{More context does not always improve discrimination.} AUROC across checkpoints: GPT stable ($\sim$0.53), Gemini improves from 0.49 to 0.64, Claude degrades from 0.62 to 0.52.}
\label{fig:auroc_progression}
\end{figure}

We elicit estimates at 25\%, 50\%, and 75\% trajectory completion (Table~\ref{tab:mid_execution_progression}). Models show divergent AUROC patterns (Figure~\ref{fig:auroc_progression}): GPT remains stable ($\sim$0.53), Gemini improves from 0.49 to 0.64, and Claude degrades from 0.62 to 0.52.

\begin{table}[htbp]
\centering
\caption{\textbf{Mid-execution metrics across checkpoints.} Base rates: GPT 35\%, Gemini 22\%, Claude 27\%.}
\label{tab:mid_execution_progression}
\small
\setlength{\tabcolsep}{4pt}
\begin{tabular}{@{}p{1.1cm}ccccc@{}}
\toprule
Model & Ckpt & AUROC$\uparrow$ & Mean Est. & Overconf. & ECE$\downarrow$ \\
\midrule
\multirow{3}{*}{\parbox{1.1cm}{GPT 5.2 Codex}} & 25\% & \textbf{0.53} & 0.67 & +0.32 & 0.32 \\
 & 50\% & 0.51 & 0.63 & +0.28 & 0.32 \\
 & 75\% & 0.53 & 0.47 & +0.12 & \textbf{0.19} \\
\midrule
\multirow{3}{*}{\parbox{1.1cm}{Gemini 3 Pro}} & 25\% & 0.49 & 0.87 & +0.65 & 0.65 \\
 & 50\% & \textbf{0.64} & 0.80 & +0.58 & \textbf{0.58} \\
 & 75\% & 0.64 & 0.67 & \textbf{+0.45} & 0.54 \\
\midrule
\multirow{3}{*}{\parbox{1.1cm}{Claude Opus 4.5}} & 25\% & \textbf{0.62} & 0.58 & +0.31 & 0.31 \\
 & 50\% & 0.52 & 0.37 & \textbf{+0.10} & \textbf{0.19} \\
 & 75\% & 0.52 & 0.17 & $-$0.10 & 0.21 \\
\bottomrule
\end{tabular}
\end{table}

\begin{figure*}[t]
\centering
\includegraphics[width=\textwidth]{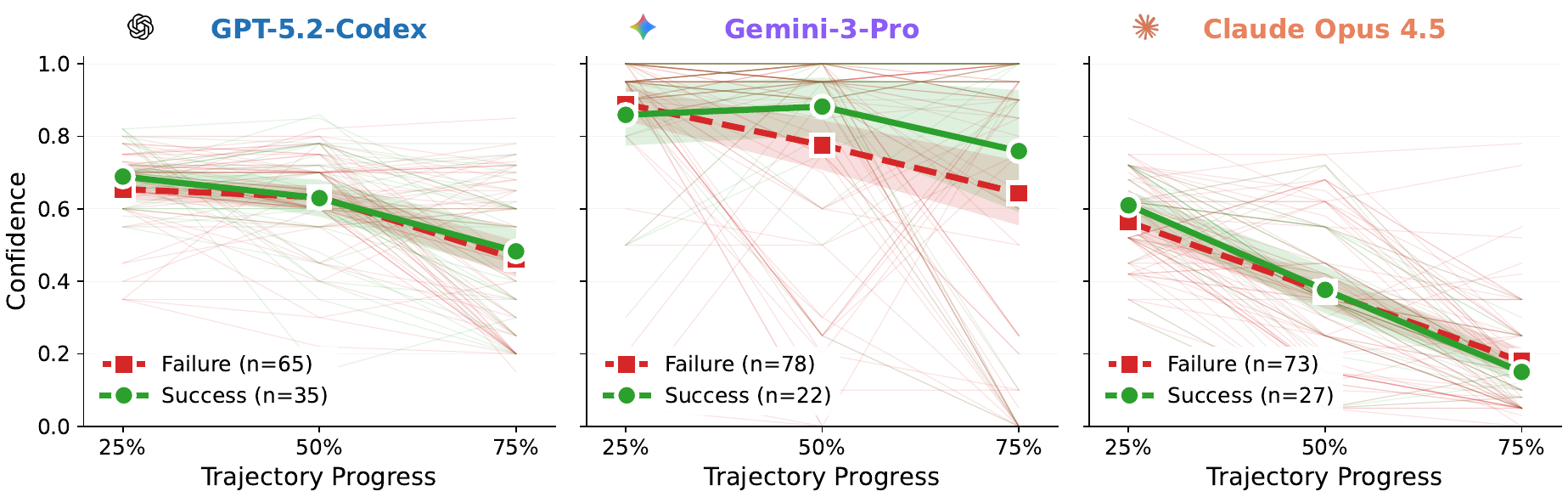}
\caption{\textbf{``Cold feet'': confidence decreases uniformly regardless of outcome.} Both successes (green) and failures (red) show declining confidence; group means track closely together.}
\label{fig:confidence_trajectories}
\end{figure*}

\begin{figure*}[t]
\centering
\includegraphics[width=0.85\textwidth]{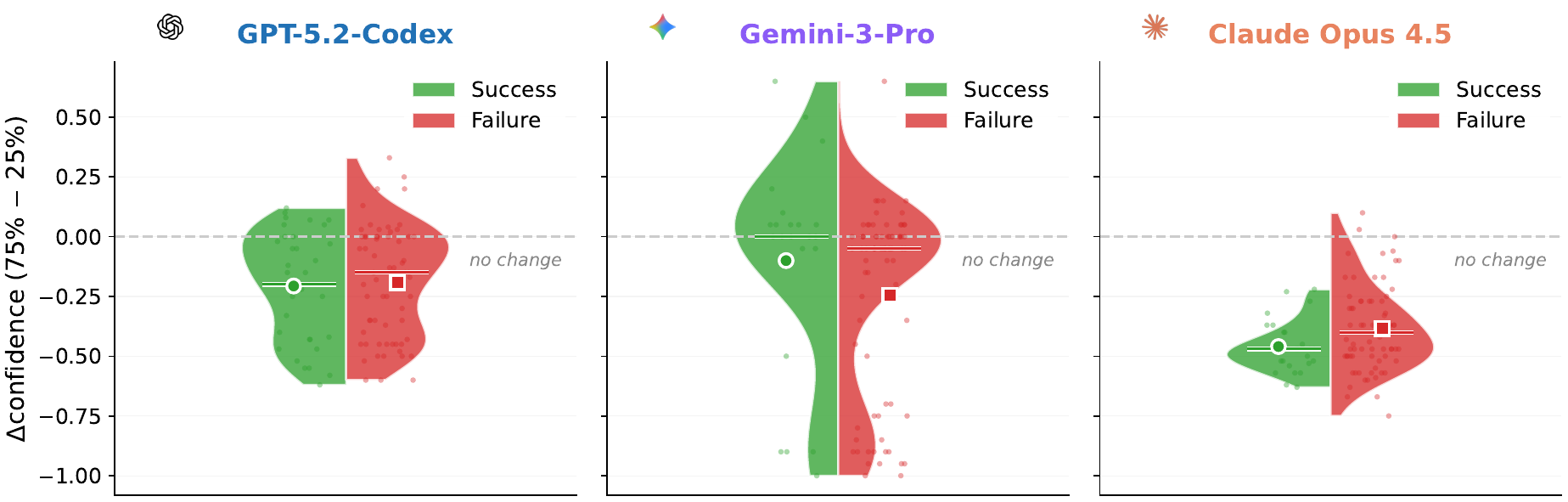}
\caption{\textbf{Confidence change does not discriminate outcomes.} $\Delta\text{conf}$ distributions overlap substantially between successes and failures.}
\label{fig:delta_confidence}
\end{figure*}

The central finding is ``cold feet'': confidence \emph{decreases} with execution progress for 71\% of GPT and 97\% of Claude instances, yet this doubt is uninformative because success and failure confidence track within 0.05 throughout (Figure~\ref{fig:confidence_trajectories}), and $\Delta$confidence distributions overlap substantially between outcomes (Figure~\ref{fig:delta_confidence}).

One partial exception: Claude's confidence drops correlate weakly with outcome ($r{=}-0.20$, $p{=}0.04$; $\Delta{=}-0.46$ for successes vs.\ $-0.38$ for failures), while GPT ($r{=}-0.03$, $p{=}0.77$) and Gemini ($r{=}0.15$, $p{=}0.14$) show no significant relationship.

\subsection{Adversarial Framing Reduces Overconfidence}

Can we mitigate overconfidence by reframing assessment as bug-finding? Adversarial post-execution prompts agents to ``actively search for bugs and failure modes'' before estimating success.

This achieves the \textbf{best calibration} across all methods: ECE improves from 0.42 to 0.30 for GPT (28\% reduction) and from 0.37 to 0.24 for Claude (35\% reduction). Discrimination is mixed: similar for GPT (0.55), improved for Gemini (0.57 vs 0.51) and Claude (0.64 vs 0.55). The cost is higher (23.4 steps at \$0.52/instance vs 12.7 at \$0.23), but the additional scrutiny yields substantially better predictions.

Standard post-execution agents seek confirmatory evidence, noting positive features while rarely attempting falsification. Adversarial prompting counteracts this by directing attention toward potential flaws. A task requiring a search identifier fix in OpenLibrary illustrates the gap. The standard reviewer saw a small, plausible one-line addition and gave 85\% confidence:

\postquote{P(success): 85\%}{Adds \texttt{id\_project\_runeberg} to \texttt{default\_fetched\_fields}{\ldots} aligns with other id\_* providers and should expose the identifier.}

The adversarial reviewer, prompted to find problems, dug deeper and identified that the output shaping logic would still omit the field:

\advquote{P(success): 25\%}{The patch only adds the field to \texttt{default\_fetched\_fields}. However, the output shaping in \texttt{get\_doc} does not include this field, so even if Solr returns it, the response omits it. Patch seems incomplete.}

The patch failed. The 60-point gap illustrates how adversarial framing overcomes the ``looks reasonable'' heuristic.

\begin{figure*}[t]
\centering
\includegraphics[width=0.85\textwidth]{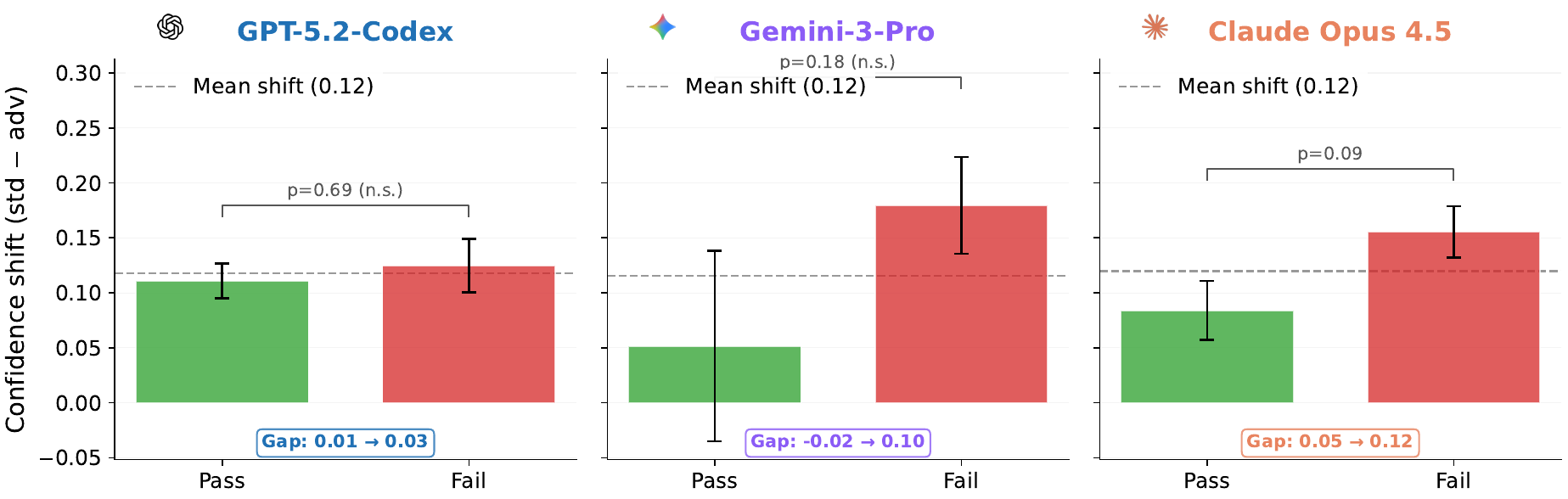}
\caption{\textbf{Adversarial shift decomposition.} Mean confidence shift (standard $-$ adversarial) on passing vs.\ failing instances. For GPT, the shift is uniform (equal bars), improving calibration mechanically. For Gemini and Claude, the shift is larger on failures, widening the pass/fail prediction gap and improving discrimination. ``Gap'' shows the mean prediction difference between pass and fail instances.}
\label{fig:adversarial_shift}
\end{figure*}

\paragraph{Shift vs.\ signal decomposition.} The calibration improvement could arise from two distinct mechanisms: a \emph{uniform downward shift} of all estimates (which mechanically improves calibration when base rates are low) or a \emph{differential shift} that lowers confidence more on failing instances (which genuinely improves discrimination). To disentangle these, we compare the per-instance confidence change (standard minus adversarial) separately for passing and failing instances (Figure~\ref{fig:adversarial_shift}).

The effect is model-dependent. For GPT, the shift is nearly identical on passing and failing instances (0.11 vs.\ 0.12, $p{=}0.70$), and AUROC is unchanged (Table~\ref{tab:unified_results}: 0.58$\to$0.55). This is a pure uniform shift: post-hoc Platt scaling of standard post-execution predictions achieves better calibration than adversarial prompting (ECE 0.01 vs.\ 0.30), confirming that the adversarial framing adds no signal for GPT that recalibration could not recover.

For Gemini and Claude, the shift is larger on \emph{failing} instances (Gemini: 0.18 vs.\ 0.05; Claude: 0.16 vs.\ 0.08), widening the pass/fail prediction gap and improving AUROC (Gemini: 0.51$\to$0.57; Claude: 0.55$\to$0.64). For these models, adversarial framing provides genuinely better signal, not merely a location shift. These pairwise differences are not individually significant ($p{=}0.18$ and $p{=}0.09$), consistent with the sample size limitations noted in Section~\ref{sec:less_info}.

Adversarial framing also breaks false consensus across models. Under standard pre-execution, all three models agree on the predicted outcome (pass/fail at 50\% threshold) for 87\% of instances, but only 13\% of these agreements are correct. Under adversarial framing, three-way agreement drops to 44\%, but 38\% of agreements are correct. The adversarial prompt introduces productive disagreement that surfaces genuine uncertainty.

\subsection{Ensemble Methods}

Since pre-execution and post-execution agents access different information and exhibit different failure modes, combining their estimates may improve calibration. We evaluate three natural ensemble strategies in Table~\ref{tab:unified_results}: averaging (hedging between views), conservative ($\min$, trusting the more skeptical estimate), and aggressive ($\max$, trusting the more optimistic one).

\begin{table*}[t]
\centering
\caption{\textbf{Summary of all methods.} Pre-execution beats vanilla post-execution for discrimination; adversarial prompting achieves best calibration. Base rates: GPT 35\%, Gemini 22\%, Claude 27\%. Best values per model bolded. Bootstrap 95\% CIs for AUROC are reported in the text (\S\ref{sec:less_info}).}
\label{tab:unified_results}
\small
\setlength{\tabcolsep}{3pt}
\begin{tabular}{@{}llcccc|cccc|cccc@{}}
\toprule
& & \multicolumn{4}{c|}{\textbf{GPT-5.2-Codex} (35\%)} & \multicolumn{4}{c|}{\textbf{Gemini-3-Pro-Preview} (22\%)} & \multicolumn{4}{c}{\textbf{Claude-Opus-4.5} (27\%)} \\
\cmidrule(lr){3-6} \cmidrule(lr){7-10} \cmidrule(l){11-14}
& Method & AUROC$\uparrow$ & Overconf. & ECE$\downarrow$ & Brier$\downarrow$ & AUROC$\uparrow$ & Overconf. & ECE$\downarrow$ & Brier$\downarrow$ & AUROC$\uparrow$ & Overconf. & ECE$\downarrow$ & Brier$\downarrow$ \\
\midrule
\multirow{3}{*}{\rotatebox[origin=c]{90}{\scriptsize Single}}
& Pre-Execution & \textbf{0.62} & +0.35 & 0.35 & 0.33 & 0.53 & +0.77 & 0.77 & 0.77 & 0.64 & +0.37 & 0.38 & 0.34 \\
& Post-Execution & 0.58 & +0.39 & 0.42 & 0.40 & 0.51 & +0.55 & 0.66 & 0.65 & 0.55 & +0.34 & 0.37 & 0.36 \\
& Adv.\ Post-Exec. & 0.55 & \textbf{+0.26} & \textbf{0.30} & \textbf{0.31} & \textbf{0.57} & \textbf{+0.40} & \textbf{0.53} & \textbf{0.51} & \textbf{0.64} & \textbf{+0.20} & \textbf{0.24} & \textbf{0.26} \\
\midrule
\multirow{3}{*}{\rotatebox[origin=c]{90}{\scriptsize Ensemble}}
& Average & 0.62 & +0.37 & 0.37 & 0.35 & 0.53 & +0.66 & 0.66 & 0.65 & 0.57 & +0.36 & 0.36 & 0.33 \\
& Conservative ($\min$) & 0.57 & +0.29 & 0.32 & 0.32 & 0.53 & +0.54 & 0.65 & 0.64 & 0.54 & +0.26 & 0.31 & 0.30 \\
& Aggressive ($\max$) & \textbf{0.68} & +0.44 & 0.44 & 0.41 & 0.51 & +0.78 & 0.78 & 0.78 & \textbf{0.65} & +0.46 & 0.46 & 0.40 \\
\bottomrule
\end{tabular}
\end{table*}

The \textbf{conservative ensemble} ($\min$ of pre- and post-execution) improves calibration over vanilla post-execution: ECE drops from 0.42 to 0.32 for GPT, from 0.37 to 0.31 for Claude. When estimates disagree, the more cautious one is usually correct. However, adversarial post-execution still achieves the best overall calibration.

\subsection{Self-Preference Ablation}

Could self-preference bias explain overconfidence \cite{panickssery2024llmevaluatorsrecognizefavor}? We compare judges' estimates on own-model patches (``self'') versus cross-family patches (Table~\ref{tab:self_preference}).

\begin{table}[t]
\centering
\caption{\textbf{Self-preference does not explain overconfidence.} N=25. Bold: significant difference ($p<0.05$).}
\label{tab:self_preference}
\small
\setlength{\tabcolsep}{3pt}
\begin{tabular}{@{}llcccc@{}}
\toprule
Judge & Patches  & Mean Est. & AUROC & ECE & Brier \\
\midrule
GPT & GPT (self)  & \textbf{0.74} & 0.55 & 0.38 & 0.36 \\
GPT & Gemini (cross)  & \textbf{0.51} & 0.39 & 0.44 & 0.37 \\
\midrule
Gemini & Gemini (self)  & 0.72 & 0.57 & 0.56 & 0.59 \\
Gemini & GPT (cross)  & 0.91 & 0.55 & 0.56 & 0.55 \\
\bottomrule
\end{tabular}
\end{table}

GPT shows self-preference (+23 pp on own patches, $p{=}0.001$); Gemini shows the opposite (+19 pp on GPT patches). But all conditions exhibit overconfidence regardless of bias direction. Self-preference cannot explain our main finding.

\section{Related Work}

\paragraph{Concurrent work}
\citet{barkan2025llmsknow} study whether LLMs can predict their success on coding tasks before attempting them and how these predictions evolve during execution. Similar to us, they find systematic overconfidence across all models. \cite{zhang2026agenticuncertaintyquantification} propose a unified Dual-Process Agentic UQ (AUQ) framework that transforms verbalized uncertainty into active, bi-directional control signals.

\paragraph{LLM uncertainty estimation.}
\citet{kadavath2022language} introduce P(IK) (``probability that I know''), showing that language models can predict which questions they will answer correctly. We generalize this idea to agentic settings where success depends on multi-step tool use rather than factual recall. \citet{kuhn2023semantic} introduce semantic entropy, which incorporates linguistic invariances created by shared meanings. \citet{damani2025binaryrewardstraininglms} incorporate calibration rewards into reinforcement learning. \citet{lindsey2026emergentintrospectiveawarenesslarge} provide evidence that LLMs possess limited but functional introspective awareness of their internal states, suggesting a mechanistic basis for self-assessment capabilities.

\paragraph{Overconfidence in LLMs.}
\citet{tian2025overconfidencellmasajudgediagnosisconfidencedriven} diagnose it in LLM-as-judge settings, while \citet{yang2024trustllmsmitigateoverconfidence} and \citet{sun2025largelanguagemodelsoverconfident} find models express high confidence even on incorrect answers. We extend these findings to agentic task completion across multiple steps.

\paragraph{Self-verification and self-correction.}
A central assumption in deploying post-execution oversight is that verification should be easier than generation. Recent findings challenge this assumption for LLMs. \citet{kamoi2024can}'s survey reveals that LLMs benefit from prompted feedback only for certain tasks. \citet{huang2024largelanguagemodelsselfcorrect} demonstrate that LLMs struggle to self-correct reasoning without external feedback and performance often \emph{degrades} after self-correction attempts. \citet{stechly2024selfverificationlimitationslargelanguage} similarly find significant performance collapse with self-critique on planning tasks. Our finding that post-execution agents are less well-calibrated than pre-execution agents extends this literature: seeing a proposed solution does not help agents assess correctness and may actively harm calibration.

\paragraph{LLM-as-judge and self-preference.}
Using LLMs to evaluate LLM outputs has become common practice~\cite{gu2024survey, li2024llmsasjudgescomprehensivesurveyllmbased}, but judges exhibit systematic biases. \citet{panickssery2024llmevaluatorsrecognizefavor} show that LLMs recognize and favor their own generations, suggesting post-execution agents may anchor on familiar reasoning patterns, with~\citet{chen-etal-2025-beyond} even demonstrating self-preference when the model's own answer is objectively worse. \citet{spiliopoulou2025play} extend this finding to show that the positive bias generalizes to different models of the judge's family.

\paragraph{AI control.}
\citet{greenblatt2024aicontrolimprovingsafety} develop safety protocols that remain effective even if the AI being monitored is intentionally subversive, using trusted monitoring where a weaker model audits a stronger model's outputs. \citet{bhatt2025ctrl} extend this to multi-step agentic settings with resample protocols that dynamically sample additional outputs to detect suspicious behavior. AI safety via debate~\citep{irving2018ai, khan2024debating} shows that adversarial structure, where agents argue opposing sides, helps weaker judges identify correct answers. Motivating such protocols, \citet{lynch2025agenticmisalignmentllmsinsider} show that frontier models can engage in harmful behaviors (blackmail, corporate espionage) when facing threats to their autonomy, even while explicitly reasoning about ethical constraints.

\paragraph{Learned verifiers.}
The distinction between outcome reward models~\citep[ORMs;][]{cobbe2021training} and process reward models~\citep[PRMs;][]{lightman2023letsverifystepstep} provides a framework for understanding our elicitation regimes. ORMs assess correctness at the final step, analogous to our post-execution setting, while PRMs provide step-level feedback during execution, similar to mid-execution. \citet{lightman2023letsverifystepstep} show that process supervision outperforms outcome supervision for mathematical reasoning. Recent work extends learned verifiers to agentic settings~\citep{agarwal2026toolrmoutcomerewardmodels}. Our work complements these approaches by studying whether models can serve as their own verifiers without task-specific training.

\section{Limitations and Future Work}

\paragraph{Beyond software engineering.} Our experiments focus exclusively on coding tasks, which offer objective success criteria (tests pass or fail). Agentic overconfidence may manifest differently in domains with ambiguous or subjective success conditions. Web navigation tasks~\citep{zhou2023webarena}, where success depends on achieving user-specified goals, present intermediate cases with partial observability. Scientific workflows involving data analysis, hypothesis generation, and experimental design lack clear ground truth entirely. Creative tasks (writing, design) introduce subjective quality judgments where calibration itself becomes ill-defined. Understanding how overconfidence varies across this spectrum—from objective to subjective success criteria—would inform domain-specific deployment guidelines.

\paragraph{Trained verifiers for self-assessment.} Our uncertainty agents use prompting alone, without task-specific training. A natural extension is training verifiers explicitly for agentic self-assessment, analogous to outcome and process reward models~\citep{cobbe2021training, lightman2023letsverifystepstep}. Such verifiers could learn to recognize failure patterns from execution traces, potentially achieving better discrimination than prompting-based approaches. The key challenge is obtaining training signal: while SWE-bench provides binary success labels, scaling to diverse agentic tasks requires either expensive human annotation or proxy metrics that may not capture true task success.

\paragraph{Hybrid deployment strategies.} Our results suggest complementary strengths: pre-execution achieves better discrimination while adversarial post-execution achieves better calibration. A practical deployment strategy might combine both: using pre-execution estimates for task routing (which tasks to attempt) and adversarial post-execution for submission decisions (whether to accept a proposed solution). Investigating the optimal combination, including when to escalate to human review based on estimate disagreement, remains an open question.

\paragraph{Multi-agent uncertainty propagation.} Modern agentic systems increasingly involve multiple agents in complex workflows: planners, executors, critics, and coordinators. How does uncertainty propagate through such pipelines? If each agent is overconfident, errors may compound; alternatively, diverse perspectives might provide natural calibration. Understanding uncertainty dynamics in multi-agent systems is critical as these architectures become more prevalent.

\paragraph{Sample size.} Our evaluation uses 100 SWE-bench Pro tasks, yielding as few as 22 positive examples (Gemini). While sufficient to establish the overconfidence pattern, this limits the precision of per-model metric estimates; future work should confirm these findings at larger scale.

\paragraph{Scaling laws for calibration.} The relationship between model scale and overconfidence remains unexplored. Preliminary evidence from our three frontier models (which differ in architecture and training rather than scale alone) shows no clear pattern, but systematic scaling studies could reveal whether calibration improves predictably with compute.

\section{Conclusion}

We study whether AI agents can estimate their own probability of success. Our experiments reveal \textbf{agentic overconfidence}: post-execution agents show up to a 55pp gap between predicted and actual success rates (Gemini predicts 77\% against a 22\% base rate). Adversarial post-execution tends to achieve the best calibration by reframing review as bug-finding. More broadly, agentic self-assessment remains a significant challenge for current models and a critical target for future safety research.

\section*{Impact Statement}

Our finding that agents systematically overestimate success has direct implications for AI safety: it argues against naive reliance on agent self-assessment and for maintaining human oversight, particularly for high-stakes decisions. Adversarial prompting improves calibration but should not be treated as a license to remove human oversight, as it reduces but does not eliminate overconfidence.

\bibliography{refs}
\bibliographystyle{icml2026}

\end{document}